%% file: manuscript-arxiv.tex
\documentclass[sigconf]{acmart}

\newcommand{\eg}[0]{\textit{e.g., }}
\newcommand{\ie}[0]{\textit{i.e., }}

\newcommand{\revision}[1]{{\color{black} #1}}
\newcommand{\minor}[1]{{\color{black} #1}}

\usepackage{comment}
\usepackage{stfloats}
\AtBeginDocument{%
  \providecommand\BibTeX{{%
    \normalfont B\kern-0.5em{\scshape i\kern-0.25em b}\kern-0.8em\TeX}}}

\copyrightyear{2024}
\acmYear{2024}
\setcopyright{acmlicensed}
\acmConference[HRI '24] {Proceedings of the 2024 ACM/IEEE International Conference on Human-Robot Interaction}{March 11--14, 2024}{Boulder, CO, USA}
\acmBooktitle{Proceedings of the 2024 ACM/IEEE International Conference on Human-Robot Interaction (HRI '24), March 11--14, 2024, Boulder, CO, USA}
\acmPrice{15.00}
\acmISBN{979-8-4007-0322-5/24/03}
\acmDOI{10.1145/3610977.3634966}
\usepackage[autostyle,german=guillemets]{csquotes}
\makeatletter

\usepackage{hyperref,quoting}
\quotingsetup{vskip=0pt,font={itshape,raggedright},rightmargin=0pt}
\usepackage{tikz}
\usepackage{cuted}
\usepackage{capt-of}
\usepackage{subcaption}
\begin{document}

  \title{Understanding Large-Language Model (LLM)-powered Human-Robot Interaction}

\author{Callie Y. Kim}
\authornote{Both authors contributed equally to this research.}
\affiliation{%
  \institution{Department of Computer Sciences University of Wisconsin--Madison}
  \country{Madison, Wisconsin, USA}
}
\email{cykim6@cs.wisc.edu}
\author{Christine P Lee}
\authornotemark[1]
\affiliation{%
  \institution{Department of Computer Sciences University of Wisconsin--Madison}
  \country{Madison, Wisconsin, USA}
}
\email{cplee5@cs.wisc.edu}
\author{Bilge Mutlu}
\affiliation{%
  \institution{Department of Computer Sciences University of Wisconsin--Madison}
  \country{Madison, Wisconsin, USA}
}
\email{bilge@cs.wisc.edu}
\renewcommand{\shortauthors}{Callie Y. Kim, Christine P. Lee, and Bilge Mutlu}

\begin{abstract}

\revision{
Large-language models (LLMs) hold significant promise in improving human-robot interaction, offering advanced conversational skills and versatility in managing diverse, open-ended user requests in various tasks and domains.
Despite the potential to transform human-robot interaction, very little is known about the distinctive design requirements for utilizing LLMs in robots, which may differ from text and voice interaction and vary by task and context.
To better understand these requirements, we conducted a user study ($n=32$) comparing an LLM-powered social robot against text- and voice-based agents, analyzing task-based requirements in conversational tasks, including \textit{choose}, \textit{generate}, \textit{execute}, and \textit{negotiate}. 
Our findings show that LLM-powered robots elevate expectations for sophisticated non-verbal cues and excel in connection-building and deliberation, but fall short in logical communication and may induce anxiety.
We provide design implications both for robots integrating LLMs and for fine-tuning LLMs for use with robots.

} 

\end{abstract}


%
%
%

%
%
%
%
%
%
%
%
%
%

%

\maketitle
\input{figure_teaser}

\section{Introduction} %

Across a wide range of day-to-day activities, robots are envisioned to possess social and communication skills that allow them to engage seamlessly and naturally with users \cite{breazeal2016social, leite2013social}. 
Past research on robots has focused on developing these skills, including conversational speech \cite{james2018artificial, laban2020tell}, gestures \cite{huang2013modeling, chidambaram2012designing, salem2012generation}, gaze \cite{luria2018effects, mutlu2009footing, mutlu2012conversational}, and appearance \cite{lee2022unboxing, hoffman2015design,luria2016designing} to facilitate effective, continuous, and dependable interactions with users.
The recent emergence of large-language models (LLMs) provides a novel opportunity for robots to augment their social and communicative abilities \cite{vemprala2023chatgpt}.
As these models enable lifelike conversations, contextual adaptation, and consistent interaction \cite{tamkin2021understanding, brown2020language}, robots can leverage these capabilities to improve their communicative proficiency to effectively address diverse user requests across a range of tasks and application domains. Despite the immense potential of LLM-equipped robots to transform human-robot interaction, a gap exists in the knowledge regarding the unique design requirements for robots that harness LLMs for conversational and communicative skills, as well as which tasks most benefit from \minor{utilizing these capabilities}.

Robots are known to have a unique effect on user experience and perceptions compared to other forms of embodiment, including text, voice, or virtual agents \revision{\cite{MUTLU2021101965, luria2017comparing, deng2019embodiment, li2015benefit, powers2007comparing, reig2022perceptions}}. Specifically, the presence of a robot triggers different cognitive activities, behaviors, or actions of a user and elicits different responses such as increased enjoyment, perceived social competence, and trust towards the robot \cite{wainer2006role, bainbridge2008effect, deng2019embodiment, reig2019leveraging, luria2019re, jung2004effects}. Therefore, it is conceivable that when users engage with robots powered by LLMs, the embodiment of the robot can shape distinct user expectations and perceptions of the sophisticated conversational system, which might have implications for how LLMs need to be specifically designed for human-robot interaction or integrated into a robot system.

\revision{The growing interest in integrating LLMs with robots necessitates a need to understand the unique design requirements of LLMs that are expected to work with robots, including design needs tailored to the tasks and contexts in which LLM-powered robots operate.} Previous design requirements for robots have been gained through exploring user perceptions regarding various task attributes and robot roles \cite{mutlu2006task, wainer2007embodiment, deng2019embodiment}. 
This exploration can similarly uncover design opportunities and optimal tasks for LLM-powered robots, shaping future guidelines for robot design and LLM development.
\revision{To understand the design requirements for utilizing LLMs for robots and identify tasks suitable for integrating LLM-powered robot agents,} we formulate the following three research questions to guide this investigation: (1) how do people perceive robots using LLMs; (2) how do people's perceptions of robots using LLMs vary across different task settings; and (3) what task contexts benefit from the embodiment of a robot when people interact with LLMs?

To address our research questions, we conducted a user study with 32 participants that compared different agent types---\textit{text}, \textit{voice}, and \textit{robot}---to better understand people's perceptions of LLM-powered robots compared to other forms of embodiment through which people interact with LLMs. Additionally, we designed four conversational tasks--\textit{execute}, \textit{generate}, \textit{negotiate}, and \textit{choose}, based on the ``task circumplex'' by \revision{\citet{mcgrath1984groups}}---to assess which tasks can benefit from LLM-powered robots. \revision{Our findings show that LLM-powered robots elicit new expectations for sophisticated non-verbal cues, and are preferred in tasks involving connection-building and deliberation between the user and the robot. Conversely, LLM-powered robots are less preferred when the LLM's rich social capabilities result in verbose responses, logical and communication errors, or induce anxiety during task interactions.} Finally, we present design recommendations for LLM-powered robots to enhance future HRI. 
We make the following contributions: 

\begin{enumerate}
    \item Compare LLM-powered agents (\ie text-based, voice-based, and social robot) to uncover unique design requirements for LLM-powered robots; 
    \item Evaluate LLM effectiveness across tasks (\ie generate, choose, negotiate, execute) to identify optimal interaction contexts with robot embodiment;
    \item Present empirical evidence on user perceptions and preferences for LLM-powered robots in diverse task settings;
    \item Provide design implications for developing LLM-powered robots and LLMs to improve future human-robot interaction.
\end{enumerate}

\section{Related Work}
\subsubsection*{Embodiment}
Embodiment plays a pivotal role in shaping how humans perceive and engage with robots. We adopt the definition of embodiment ``structural coupling'' from \citet{ziemke2013s} such that a system is embodied if mutual perturbative channels exist. We focus on physically embodied robots that can leverage rich channels of communication such as gesture, posture, gaze, facial expressions, proxemics, and social touch \cite{deng2019embodiment}. Prior research shows that interactions with physically embodied robots lead to higher user engagement, enjoyment, trust, and empathy compared to text, voice-based, or virtual agents \cite{10.1145/1518701.1519021, bainbridge2011benefits, bickmore2005establishing, 10.1145/2696454.2696471}. Additionally, embodiment influences user behavior, affecting interaction duration and distance \cite{RODRIGUEZLIZUNDIA201583, 10.1145/1957656.1957786}. Several studies have explored how physical embodiment affects task performance and impression by comparing physically embodied robots to virtual agents \revision{\cite{wainer2007embodiment, HOFFMANN2013763, segura2012you, MUTLU2021101965}}. These studies indicate that user preferences for embodied agents are influenced not only by embodiment but also by the specific task context.

\subsubsection*{LLM in Robotics}

Robots function as the vital bridge connecting the tangible real world and LLMs. This connection enables LLM to infer knowledge from the physical environment through data collected by \minor{sensors}. Simultaneously, LLMs empower the robot with the capability to comprehend semantic meanings and engage in flexible dialogue interactions. Thus, LLMs with robots find their primary applications in task planning \cite{ahn2022can, singh2023progprompt, driess2023palme} or human-robot collaboration \cite{kodur2023structured, 10141597}. For instance, \citet{10141597} investigated the implications of LLM-powered robots when users controlled the robot through text for assembly tasks in virtual reality. 

\revision{
Researchers have also explored the effectiveness of LLMs for conversational robots in specific tasks. \citet{cherakara2023furchat} designed a system in which the robot displays appropriate facial expressions when conveying information about the National Robotarium. \citet{PPR:PPR655990} utilized LLMs to create a personalized companion robot and examined the challenges associated with open-domain dialogue when interacting with older adults. \citet{10.1145/3568294.3580067} applied LLMs to a social robot to enhance the well-being of older adults by generating empathetic responses.
\citet{yamazaki2023building} constructed a scenario-based dialogue system for a robot and demonstrated the effectiveness of LLMs while establishing trust with users. 
While prior research has primarily concentrated on evaluating the efficacy of LLM-powered robots in specific tasks, we aim to explore a wider array of tasks and contexts where LLM-powered robots can offer advantages and comprehend the unique design requirements to effectively incorporate LLMs with robots across diverse task settings.
}

\section{Method}

\subsection{Embodiment Design}
To understand people's perceptions of robots when powered by LLMs, we compare a social robot agent against two other agents--a text-based agent and a voice-based agent. All three agents were equipped with GPT-3.5, OpenAI's text-davinci-003 model \cite{brown2020language} \revision{without fine-tuning}. The model parameters were set to temperature = $0.7$ with max tokens = $2048$. \revision{Pre-prompts were used to outline the four tasks, with parameters identical to those used by \citet{billing2023language} in Pepperchat.}

\subsubsection{Text Agent}
Resembling a chatbot, the users interacted with the text agent through text input and output. The user sent and received prompts via the GPT model with OpenAI API.

\subsubsection{Voice Agent}
Simulating a voice assistant, the voice agent communicated exclusively through voice commands. For the voice agent, the participant and the robot were separated by a screen such that the participant only interacted with the agent through voice. It utilized the robot's module, ``ALAudioDevice \cite{aldebaran2023audio}'' to capture the user's speech. The audio recording is then sent to Google Cloud service \cite{google2023speech} for speech-to-text analysis, then forwarded to the GPT model via OpenAI API. The GPT model generates a response, which is converted into a speech using the robot \minor{Pepper's \cite{pepper2023}} module, ``ALAnimatedSpeech \cite{aldebaran2023animated}.'' 
\revision{The same robot was used for both the voice and robot agent instead of a smart speaker to avoid favoring one specific technology over another within the broad space of voice-based agents (\ie smart speakers, smart displays, and virtual assistants) and to ensure consistent voice interactions across both voice and robot agent conditions.}

\subsubsection{Robot Agent}
The social robot, Pepper, was used to engage with users through animated gestures, text-to-speech, and face recognition. For successful communication between the participant and the LLM-powered robot, we employed Pepperchat \cite{billing2023language}, which utilizes Google Cloud speech-to-text functionality for speech-based dialogue, contributing to a seamless and responsive communication experience. 
\revision{
We chose a minimalist design for the robotic agent, emphasizing its basic embodiment to highlight high-level differences among text, voice, and robot embodiments, rather than fully utilizing non-verbal cues.
Thus, we chose to accept an out-of-the-box implementation of each agent, rather than each agent having specific design features (\eg visual cues for the voice agent.)}

\input{figure_method}

\subsection{Task Design} %
To understand the design requirements for LLM-powered robots across various task settings, we designed different tasks based on the Group Task Circumplex Model proposed by \revision{\citet{mcgrath1984groups}}. The circumplex model is structured around two dimensions, ranging from conflict-based to cooperative, and conceptual to behavioral. The circumplex model classifies group tasks into four categories: (1) generate: tasks that involve generating ideas or plans; (2) choose: tasks that involve choosing a solution or plan from a set of alternatives where the correct or agreed-upon answer exists; (3) negotiate: tasks that involve resolving conflict of viewpoints, interests, and motives; and (4) execute: tasks that involve executing a plan or performance. This framework offers a structured approach to comprehend the nature of the tasks that groups undertake. Figure \ref{fig:method} shows examples of task interactions. Below we discuss the specific tasks designed for our study.

\subsubsection{Generation Task}
In the generation task, the agent and the participant collaboratively create an imaginary story. Participants were asked to follow a general guideline to introduce characters, features of the characters, and the setting for story development. To create the foundation and actual story, the participant and agent took turns each adding a sentence. To construct a comprehensive story, the participants were told to ideally incorporate obstacles, solutions to address the obstacles, a climax in the story, and a plot.

\subsubsection{Choosing Task}
In the choosing task, the agent assisted the participants in selecting a subset of items from a collection of items. There was a different theme for the collection of items for each task, including a ski, beach, and camping trip. Participants were told to select items that focused on practicality over leisure. The item criteria were based on those commonly featured as essential on various travel websites. Participants engaged in discussion with the agent to finalize their item list.

\subsubsection{Execution Task}
In the execution task, the agent acted as an instructor and the participant acted as a student. The agent's role was to teach the participant how to prepare a beverage in a cafe setting. Only the agent knew which drink to make and participants were asked to follow the instructions. Participants were told to ask the agent if they had any confusion or questions. %

\subsubsection{Negotiate Task}
In the negotiation task, the agent acted as a seller of second-hand items and the participant acted as the potential buyer. The agent's goal was to sell the item as expensive as possible and the participant's goal was to buy the item as cheap as possible. The agent was not aware of how much money the participant held. To control the task settings and provide consistency, an absolute minimum price line was set for the item. %

\section{User Study}
\subsection{Study Design}
The study followed a mixed-factorial design with scenario tasks as the between-subjects factor and the agent embodiment as the within-subjects factor. Participants were randomly assigned to one of four tasks (\ie generate, choose, execute, and negotiate) and then engaged with the three different agents (\ie text agent, voice agent, and robot) in counterbalanced order.
At the beginning of the study, participants were shown interaction examples with the LLM-powered agents that involved disagreeing with suggestions, asking follow-up questions, and tracking task progress.
Additionally, the task given per agent differed slightly in topic to avoid the learning effect (\eg a camping, beach, and ski trip). \revision{Prompts for the tasks can be found in the supplementary materials}\footnote{The supplementary materials can be found at \url{https://osf.io/exjrd/?view_only=88c0b1ff4b2b4f969928a614c9fa8fff}}. After interaction with each agent, participants completed questionnaires and a semi-structured interview about their experience. All sessions were held in person, 
\minor{audio and video recorded through Zoom \cite{zoom2023} on a laptop}.

\subsection{Measures}
\subsubsection{Subjective Measures}
To measure participants' perception of the agents, we used a modified version of the Godspeed questionnaire \cite{bartneck_measurement_2009}, which includes a series of semantic scales for measuring the robot's animacy (Cronbach's $\alpha = 0.91$), anthropomorphism (Cronbach's $\alpha = 0.89$), likeability (Cronbach's $\alpha = 0.93$), perceived intelligence (Cronbach's $\alpha = 0.91$), and perceived safety (Cronbach's $\alpha = 0.72$) on a five-point rating scale. We modified the questions such that the items asked about their perceptions of ``agents'' instead of ``robots.'' 
Our analysis of the item reliability of the perceived safety subscale found a Cronbach's $\alpha$ of $-0.27$, due to a miscoded item in the subscale. 

Godspeed items are written in a consistent way, such that in each group high values of a variable indicate a similar direction. Specifically, the miscoded variable, \textit{still} (anchored at 1) to \textit{surprised} (anchored at 5) appeared to differ in direction from other items: \textit{anxious} (anchored at 1) to \textit{relaxed} (anchored at 5) and \textit{agitated} (anchored at 1) to \textit{calm} (anchored at 5). After re-coding the \textit{still-surprised} item by flipping the scale so the semantic meaning of the item would be consistent with others, we calculated Cronbach's $\alpha$ of $0.72$. This miscoding of the \textit{still-surprised} item and correction by reverse-coding have been reported by prior work that used the Godspeed questionnaire \cite[\eg][]{schillaci2013evaluating, 10.1145/3029798.3038394, Bodiroža2017Gestures}. Additionally, upon closer inspection of the items of this subscale, which included anxious-relaxed, agitated-calm, and surprised-still (after reversion), we determined these items to be a poor fit to the overall construct of ``perceived safety'' and decided to exclude it from our analysis.

In addition to the Godspeed questionnaire, we measured participants' satisfaction (Cronbach's $\alpha = 0.96$) with the interaction on a seven-point rating scale (1 = strongly disagree; 7 = strongly agree) using the satisfaction subscales from the Usefulness, Satisfaction, and Ease of Use (USE) Questionnaire proposed by \citet{lund2001measuring}. The overall Cronbach's $\alpha$ value for the Godspeed attributes and satisfaction was 0.97.

\subsubsection{Behavior Measures}
To observe and understand participant behaviors, \minor{we collected measures of the total number of input tokens derived from the participants' prompts.} This approach involves counting discrete units that the OpenAI API divides from the user's input to process the prompt. This metric enables assessment of the length of \minor{dialogue input provided by the user within the} conversation during the task.
\subsubsection{Performance Measures}
To understand the quality of the interaction, we measured the number of failures that occurred during the interaction. 
\revision{
We considered two categories of failures: (1) \textit{technical errors}, such as interruptions by the agent, and inaccurate transcriptions from Automatic Speech Recognition (ASR); and (2) \textit{hallucinations}, where the response from the LLM is nonsensical or unfaithful to the provided source input \cite{10.1145/3571730}.
}

\subsection{Participants}
We recruited 32 participants (10 male, 20 female, 1 gender-queer, 1 non-binary) through a university mailing list between the ages of 18 and 59 ($M = 27.47, STD = 10.30$) where 69\% were White, 28\% were Asian, and 3\% preferred not to answer. Participants were required to be in the United States, fluent in English, and at least 18 years old. All participants agreed to participate in our study via our institution's IRB-approved consent form. The study lasted for approximately 60 minutes and participants received \$15 per hour for compensation upon study completion.

\subsection{Analysis}
Factorial repeated-measures analysis of variance (ANOVA) was used to determine whether the task and agent embodiment had a significant effect on all measures. 
\revision{If the ANOVA test showed significant effects, we tested our data for pairwise differences using Tukey honest significance test (HSD), which controls for Type I error considering all possible comparisons.} The qualitative data was analyzed using Thematic Analysis (TA), following the guidelines developed by \citet{clarke2014thematic} and \citet{McDonald19}. The first authors became acquainted with the data by conducting the studies and initially creating a codebook \cite{decuir2011developing}. Through ongoing team discussions, codes were grouped into categories and refined until a consensus was reached. These categories were then further organized and reiterated to extract themes that emerged from our study data. Once all potential themes were reviewed, the final themes are presented as our findings.

\section{Results}

We present the findings derived from our quantitative and qualitative data analysis. In section \ref{sec:finding_1}, we show the results of our quantitative data analysis highlighting the overall patterns from the interactions between the LLM-powered agents and participants. As the quantitative data showed high variance, we present the findings of our qualitative analysis in Section \ref{sec:finding_2} to Section \ref{sec:finding_5}, to gain further insights into the detailed factors that affected user preference and perceptions towards LLM-powered robots. 

\input{test}

\subsection{Data from Quantitative Measures}\label{sec:finding_1}
We examined the influence of embodiment on interactions with LLM-powered agents through an analysis of data from our quantitative measures. Figure \ref{fig:test} summarizes significant findings. 
\revision{Overall, embodiment had a significant effect on input prompt length, $F(2, 56) = 14.30, p < .001$.} 
When comparing the input length across embodiment conditions, participants provided significantly longer inputs to the text agent than the voice agent or the robot. 
\revision{Embodiment also had a significant effect on input length within tasks, $F(6, 56) = 4.25, p = .001$.}
The generation task, in particular, had a significantly longer length of input in the text condition than other embodiment conditions. 
\revision{Finally, embodiment had a significant effect on failures, $F(2, 56) = 55.16, p < .001$.} \minor{In comparing failures across embodiment conditions, participants encountered the most failures with the voice agent, followed by the robot and text agents.}
We observed a higher occurrence of failures in the generation task, underscoring the difficulties faced by agents that used voice-based input when confronted with extended input, especially within this specific task context, \revision{$F(6, 56) = 5.94, p < .001$}. 
The diverse range of user experiences related to the quality of interactions led to a significant variance in participants' satisfaction when interacting with different agents, \revision{$F(2, 56) = 3.81, p = .028$}. Participants rated their level of satisfaction with the text agent to be higher than the voice agent and marginally higher than the robotic agent. We attribute these differences in the satisfaction scores to the results of the failures participants experienced with voice-based interaction with the voice and robotic agents. There were no statistically significant differences across embodiments or tasks in other subjective metrics, including anthropomorphism, animacy, likeability, and perceived intelligence, \revision{which can be found in the supplementary materials.}

\subsection{Data from the Qualitative Measures}\label{sec:finding_2}

In this section, we present the findings of our qualitative analysis in the order of tasks in which LLM-powered robots were more preferred by participants, namely: (1) execute; (2) negotiate; (3) choose; and (4) generate.
In each task category, we present design themes explaining the positive and negative effects of LLM-powered robot agents, supported by quantitative results.
\subsubsection{Execute} Below are themes that emerged in the Execute task.

\paragraph{Conversational Interactions for Effective Learning}
Across all the agents, the LLM's capability to facilitate natural conversations while delivering instructions and responses with contextual understanding significantly benefited the participants' engagement in the interaction. For the execution task, the agent received task instructions before engaging with the participant and then responded freely to the participant's requests. All participants frequently sought guidance on how to proceed in the task, thereby leading to concise and clear prompts that were easy for the agent to comprehend and respond to. As shown in Figure \ref{fig:test}, the input length of the prompts tended to be shorter in the execution task. Moreover, \revision{seven} participants expressed satisfaction with the agent's response, the LLM's contextual understanding ability enabled the agent to provide sufficient responses to follow-up questions. \textit{P26: ``The robot was able to answer all the spontaneous questions that I had for it, which really surprised me and we were able to have an actual conversation. He's smart enough to teach me!''} Given the seamless communication, there were minimal instances of agents failing to understand and respond logically to requests, as shown in Figure \ref{fig:test}.

\paragraph{Robot's Social Aspects Enhancing User Engagement}

\revision{Six} participants expressed a preference for the robot agent over the voice and text agents due to its efficiency in interacting and enriching engagement with the social aspects of the robot. As participants physically prepared drinks while simultaneously seeking instructions from the agent, they expressed that interacting with the robot or voice agent through spoken communication was easier and facilitated multitasking, unlike the text agent, which required them to pause their actions and type queries. \textit{P25: ``I could start asking the follow-up question as I was doing a task versus text, I had to finish the whole task and then type the question. I thought it went by a little smoother.''} \revision{Five} participants encountered additional difficulties with the voice agent, struggling to time their prompts with the voice agent, leading to discomfort and reduced interest in engaging with the agent. \textit{P30: ``That one [voice agent] for me feels the most choppy and disconnected, so it was hard for me to tell when I could ask something compared to the others [agents].''} %

Moreover, \revision{four} participants noted that the robot's social cues and physical presence enhanced their receptiveness to instructions and task engagement, as it resembled real-life communication. \textit{P26: ''When you're able to see Pepper, you can kind of look at the tilted head to understand whether it's like thinking or not. But when you can't see Pepper, it's like, what's going on? Those little things help us communicate.''}  \revision{Four} participants also expressed appreciation towards the robot's social cues, such as maintaining eye contact, waiting for task completion, and offering encouragement, as these interactions made participants feel a genuine sense of companionship and support. \textit{P27: ``So especially when you're learning, part of the learning is from interacting. And that relates to the emotional connections and things that are underneath. So you need the actual robot to do it together, physically engaging.''} The social cues presented with the robot's social presence increased the participants' focus and immersion in the task, driven by a desire to \textit{P25: ``impress the robot, because he is watching me.''} Finally for future interactions, \revision{five} participants envisioned the robot utilizing its arms and body parts for instruction demonstration. \revision{Participants explained that they expected the robot to have sophisticated non-verbal cues to match the advanced capabilities of its conversational skills. 
\textit{P26: ``The robot's movements reminded me it was still in development. They were random and didn't have any relation to what it was saying, when the way it talked was such high quality. Made it kind of creepy.''}}

\input{figure_findings}

\subsubsection{Negotiate} We found the themes below in the Negotiate task.\label{sec:finding_3}

\paragraph{Information Exchange with Contextual Understanding}
During the negotiation task, participants engaged with the agent to reach a mutual agreement on the price of an item. This negotiation process involved participants posing questions about item specifics, usage history, potential bundle deals, and more. The LLM's ability to understand the context within a conversation, considering the dialogue history to generate coherent responses, was effective in maintaining a seamless, life-like conversation. \textit{P22: ``Okay, hold on [robot], first let's sit down and talk about this more. Tell me a little bit more about this bike. Once all my questions are answered sufficiently, then we can start to negotiate.''}

Participants' queries for negotiation were generally longer when interacting with the text agent compared to the voice and robot agent, as shown in Figure \ref{fig:test}. \revision{Five} participants explained that it was more convenient to input their questions via text to the agent as opposed to verbally articulating their inquiries. The primary questions posed across the agents were largely similar, with participants posing additional queries based on the agent's response. The text and robot agents had clear distinction when their response was finished, as the text agent displayed a complete response on the screen, while the robot agent indicated completion through non-verbal cues such as putting down its arms, tilting its head, and adjusting its gaze. As illustrated in Figure \ref{fig:test}, participants encountered difficulties less frequently when engaging with these agents, whereas the voice agent encountered a higher incidence of failures, as participants had challenges in determining when to speak and whether the agent accurately understood their prompts.

\paragraph{Robot Establishing Rapport for Negotiation}
Although the robot agent was not the most convenient to use, \revision{four} participants expressed that the robot was most effective in establishing the connection and rapport that was required for successful negotiations. \revision{Two} participants described that building rapport and personal connections with the agent was crucial to resolving the conflicts for negotiation. \textit{P22: ``Negotiation starts with building trust, obviously.''} The social characteristics, such as the natural language produced by the LLM and the behavioral aspects of gaze, facial expressions, and body movements, contributed to a sense of engaging in a genuine conversation with a social entity. \revision{Five} participants described that the ability to see and physically interact with the robot agent created a personal interaction atmosphere, enhancing the agent's reliability compared to the voice or text agent that lacked physical social cues. \textit{P17: ``I do feel like for me that's important. Just to be able to engage with some visual cues, eyes, you know, a face, that seems to be more appealing, inviting a further engagement. Making a real conversation.''}
Although the text agent was efficient in providing immediate and informative responses, it was perceived more as a search engine and less as an agent genuinely interested in mediating deals to participants' preferences. This perception made \revision{three} participants less inclined to negotiate for more expensive items with the text agent. \textit{P5: ``Yeah coffee machine no big deal, but the car I wouldn't just negotiate that with the text [agent]. It seems too machinery and sketchy.''} Similar to other tasks, the voice agent was least positively perceived among agents for negotiations, as it was challenging to discern intentions and gauge conversational progress during the negotiation. 

\subsubsection{Choose} We identified the themes below in the Choose task. \label{sec:finding_4}

\paragraph{Recurrence of Errors in Communication and Logic}
During the choosing task, participants chose a final set of items from a list of items based on practicality and preference through discussions with the agents. The discussions required led participants to articulate the reasons for and validate their selections to the agent. Participants also disagreed with the agent's suggestions, prompting them to elaborate on their rationale. Similar to the negotiation task, \revision{five} participants described that the text agent was the most effective at facilitating accurate and expeditious information exchange compared to the voice and robot agents. This efficiency resulted in longer prompts in the text agent, as illustrated in Figure \ref{fig:test}. Additionally, as participants iterated the item selection criteria and requested validation on the final item selections, the LLM occasionally exhibited errors in logic or inconsistencies with the dialogue history. For instance, one participant described \textit{P24: ``He [robot] also had some unusual answers. I asked what I should bring on my ski trip and Pepper said that I could bring a sand baking tray and ski on the tray. I was like, not sure how that would work out. So I felt less engaged in obviously,  I felt less of a connection with Pepper in this instance.''} In another example, the agent altered its recommendations multiple times or presented conflicting arguments for a single item. Unforeseen failures, such as misinterpreting participants' requests, prematurely responding before participants, and introducing flawed logic in the agent's responses, led to increased failures of the voice and robot agent as illustrated in Figure \ref{fig:test}. \textit{P10: ``There's often a disconnect between, it [agent] knowing the facts but not knowing that it doesn't make sense.''} Such failures were further described by \revision{four} participants to decrease the satisfaction and motivation to engage with the voice or robot agent.

\paragraph{Inefficient and Time-consuming Interaction with Robot}
During the task, participants described that they initially held a general idea of the items they intended to select, and they intended to have specific discussions with the agents to efficiently narrow down their choices. Participants sought details regarding the advantages and disadvantages of these items and validation to determine their inclusion. During this engagement, \revision{four} participants noted that the conversational interactions necessitated repetitive and overly verbose exchanges with the agent to acquire information equivalent to a ``quick search,'' resulting in an undue amount of time. In contrast, the text-based agent promptly provided the requested information and additionally preserved logs for users to reference when finalizing their decision. \textit{P8: ``I appreciated Pepper being all nice, but sometimes it wasn't the exact information I was looking for and there was a lot of fluff. And then I would have to wait for him to finish to ask again. And in the end, I don’t even remember what he said! So in those terms, the text was much more efficient.''}

\subsubsection{Generate} The Generate task included the themes below.\label{sec:finding_5}

\paragraph{Communication Barriers for Creative Collaboration}
The creative nature of the generation task guided participants to devise prompts that were more personal and situation-specific. These prompts included the introduction of character names and attributes, intricate plot developments, and distinctive settings. As a result, participants' prompts tended to be more extensive during the generation task, as shown in Figure \ref{fig:test}. A difference in input length appeared between the text agent versus the voice and robot agents, as participants expressed difficulties in verbally expressing their creative thoughts.

These communication difficulties were due to the timely manner of the task. 
Participants frequently encountered situations where they had a lot to convey but struggled to do so spontaneously in real-time, without excessive pauses or verbosity in response to the robot. Moreover, the collaborative nature of this creative process meant that participants needed additional time to carefully consider how they wanted to incorporate the agent's ideas into their story and shape the subsequent storyline. As a result, \revision{seven} participants found verbal communication to be less preferable and more challenging compared to using text inputs, where text inputs allowed them to express their ideas in a more organized manner. The communication difficulties led to malfunctions as the agents frequently misinterpreted the user's prompts, overlooked important elements of the participants' instructions, or interrupted the participants. \textit{P7: ``But for both cases, the voice and the robot agent. There are a couple of times that they just ignore, or interrupt what you say that makes you more frustrated. It would just start rattling off when I wasn't done talking.''} Among all the tasks, the generation task showed the highest number of communication failures, as shown in Figure \ref{fig:test}. Thus, \revision{six} participants perceived the text agent to be more practical in tracking the storyline and avoiding communication errors.

\paragraph{Discomfort with Robot in Content Creation}

During the task, \revision{four} participants expressed discomfort with the robot's social presence when trying to contemplate creative ideas. This discomfort was due to the participants' expectations of the agent being ``smart,'' due to its sophisticated verbal capabilities from the LLM. \textit{P19: ``It seems like I am talking to a person, because it [robot] is so smart, and I'm like, oh, they might remember this.''} These expectations made the robot's social presence cause pressure on the participants, and even anxiety when they were trying to come up with the next storyline. As the robot would continue to gaze at the user or make subtle movements and facial expressions as it awaited the next prompt, the participants described that this action created a sense of urgency, compelling them to generate their ideas quickly without allowing for thorough reflection. \textit{P6: ``I thought more when I was using the text, because it wasn't just off the top of my head. But with the robot, I did just kind of say more random stuff because I felt like I needed to respond right away.''} Another participant supported this finding by describing \textit{P19: ``I felt the most comfortable with the text event in generating ideas because I had no accountability. Or the robot looking at me in the face. I feel a little bit more embarrassed.''}

\section{Discussion}

\revision{In this work, we explored the distinctive design requirements for integrating LLMs with robots. To understand how LLMs should be tailored for robot applications, we conducted a user study involving 32 participants that compared a text, voice, and robot agent across four tasks: execute, negotiate, choose, and generate. 
Our findings show that the LLM-powered robot elicited user expectations for sophisticated non-verbal cues and was more favored in the Execute and Negotiate tasks, where building connections and engaging in social discussions were crucial. However, LLM-powered robots were less preferred in the Choose and Generate tasks, due to communication difficulties and the potential anxiety during collaboration.
Below, we present design implications that address the distinctive design needs for robots utilizing LLMs, as well as the unique design requirements for LLMs intended for use with robots.}

\subsection{Combining LLM-powered Robots with Non-verbal Interaction Cues}
\revision{Our findings reveal that interactions with LLM-powered robots established unique expectations for users regarding non-verbal cues. In contrast, users interacting with text and voice-based agents did not actively seek non-verbal cues. These expectations were not solely shaped by the robot's physical form but were rather a result of the robot's advanced language capabilities powered by the LLM. This sophistication in language abilities led users to anticipate equally sophisticated non-verbal cues from the robot. Therefore, it is recommended that LLM-powered robots explore and incorporate a diverse range of rich non-verbal cues, such as gaze \cite{PPR:PPR655990}, gestures \cite{lee2023developing, 10.1145/3586182.3616623}, behaviors, and facial expressions \cite{cherakara2023furchat}, during interactions with users. These non-verbal elements should be tailored to match the heightened expectations set by the robot's advanced spoken language capabilities. For instance, combinations of non-verbal cues can be developed to demonstrate appropriate behaviors and explanations in various contexts, such as reactions to different user inputs, respecting user boundaries, and failures in responding to user requests. The alignment of verbal and non-verbal cues can enhance the experience between the user and LLM-powered robots, making the engagement more sophisticated and natural for users.}

\subsection{Considering Task Characteristics when Utilizing LLMs with Robots}
\revision{Our study findings indicate that LLM-powered robots exhibited a preference for certain tasks over others due to the unique characteristics of each task. 
Therefore to effectively utilize LLMs for robots, customization \cite{yamazaki2023building, 10.1007/978-3-031-48306-6_15} and fine-tuning \cite{10.1007/978-3-031-45673-2_46} are crucial for different tasks. While existing state-of-the-art LLMs can be suitable for tasks similar to Execute and Negotiate, for tasks resembling Choose and Generate, LLM adaptation will be required. One method can be fine-tuning LLMs to fit the goal and context of the task, such as simplifying rich social descriptions to enhance efficiency, intuitiveness, and directness. The fine-tuning process can include selecting a pre-trained model, defining task objectives \cite{ge2023openagi}, preparing task-specific data \cite{10.1145/3568294.3580067}, configuring fine-tuning parameters \cite{pmlr-v97-houlsby19a, DBLP:journals/corr/abs-2106-09685, liu2022few}, training the model, validating and evaluating performance \cite{peng2023check}, and deploying the fine-tuned model onto the robot.}

\subsection{Utilizing LLMs for Robot Design}
\revision{During our study, we observed several design opportunities to leverage LLMs with robots. During the interactions with users, LLMs demonstrated the potential to empower robots to adapt to a broader array of user requests and effectively capture user needs and preferences. These instances emphasize the capacity of LLMs to either substitute or complement traditionally challenging tasks in the realm of robot design and implementation. 
For instance, during robot application development, significant time is often spent implementing the dialogue system, defining the robot's intent and entity, and training it to handle user requests and communication variability effectively.
LLMs can address these challenges by flexibly accommodating task models variations and processing a wide range of inputs. This adaptability can guide robots to offer personalized user experiences through iterative and engaging interactions.

However, it is crucial to acknowledge that integrating LLMs may also introduce risks and errors, such as causing robots to deviate from context or produce hallucination errors. %
LLM-powered robots in real-world settings may display unexpected behaviors or make statements inconsistent with their intended character, leading to a mismatch between the situational context and the robot's intended personality.
Furthermore, as illustrated by instances in our study, LLMs may introduce hallucination errors, leading the robot to provide information that is inaccurate or nonsensical. As a result, LLMs on robots must be viewed as both a feature and a potential challenge, requiring the establishment of appropriate boundaries regarding what LLMs can and cannot achieve. Technical methods such as curated datasets for pre-training \cite{wang2023selfinstruct, taori2023stanford}, program verification \cite{shen2021generate, cobbe2021training}, human-in-the-loop review \cite{ouyang2022training}, fine-tuning, and other measures can be used for LLM action boundaries. }

\subsection{Limitations and Future Work}
Our study has several limitations. First, we chose to compare LLM-powered robots to two other forms in which people interact with LLMs: text agents and voice agents. \revision{While this comparison was informative on how people's perceptions of LLM-powered robots differed from other LLM-powered agents, we would have ideally compared the LLM-powered robot to a non-LLM-powered robot. However, it was difficult to specify} what a ``non-LLM'' condition would look like and how such a condition would be implemented. Nonetheless, the lack of comparison against a non-LLM-powered robot limits our ability to study the unique effects of the integration of LLMs in robots. We plan to explore this question in our future work, for example, using a Wizard-of-Oz approach \revision{with human operators generating responses or a rule-based approach with} scripted dialogues to achieve sophisticated but fixed conversational capabilities for the robot. 
Second, the quantitative data from subjective measures exhibited high variance, leading to non-significant results in multiple items. 
\revision{
This high variability can be attributed to our sample size, which limits the generalizability of our findings. Future work may include larger-scale studies. 
Third, our minimalist robot design lacked diverse non-verbal behavior, potentially causing users to perceive the three agents as more similar than in real-world scenarios.
Future research can explore how LLM-powered robots might use the full range of their embodied capabilities, which could also improve their communication performance with users. %
}

\section{Conclusion}
\revision{This research investigates the design requirements for robots connected to LLMs and in tasks where they excel. We compare three LLM-powered agents---text, voice, and robot---across four tasks---generate, negotiate, choose, and execute. Findings reveal that LLM-equipped robots enhance user expectations for non-verbal cues, excel in connection building and deliberation, but face challenges in communication difficulties and creating social pressure. We provide design insights for robots adopting LLMs and LLMs used for robots.}

\begin{acks}
This work was supported by the Sheldon B. and Marianne S. Lubar Professorship, an H.I. Romnes Faculty Fellowship, and the National Science Foundation award (\#1925043).
\end{acks}

\balance


%
%
%
%
%
%
%
%
%
%
%
%
%
%
%
%
%
%
%
%
%

%
%
%
%
%
%
%
%
%
%
%
%

%
%
%
%

%
%
%
%
%
%
%
%
%
%
%

%

%

%
%
%
%
%
%
%
%
%
%
%
%
%
%
%
%
%

\end{document}

%% file: figure_teaser.tex
\begin{figure}[t]
\centering
  \includegraphics[height=3in]{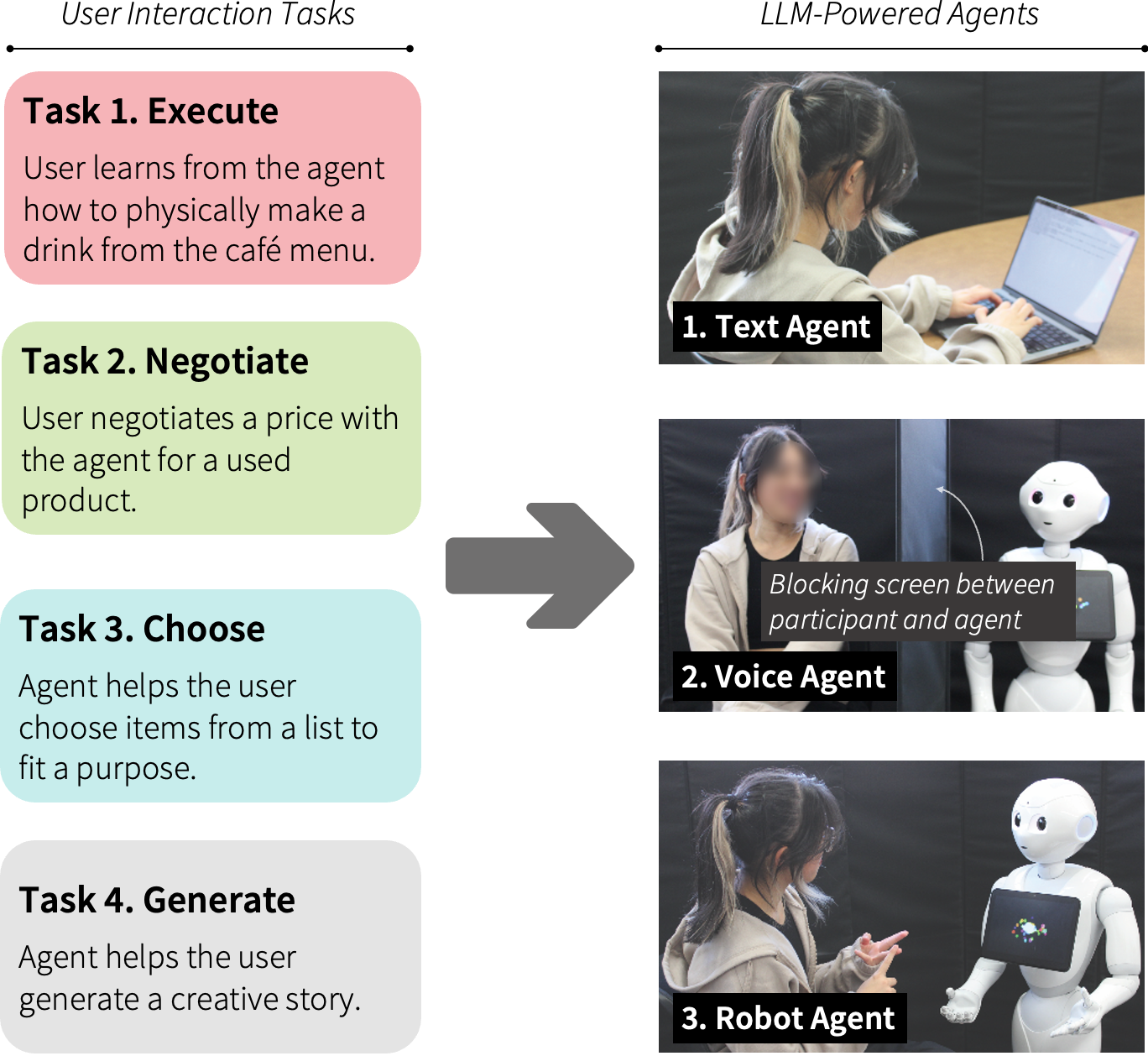}
  \caption{We investigate people's perceptions of and preferences toward LLM-powered robots. We conducted a user study that compared an LLM-powered social robot against text-based and voice-based agents. \textit{Left:} Users participated in one of four tasks: choose, generate, execute, and negotiate. \textit{Right:} The user engages with (1) the text-based agent by entering and receiving text-based prompts, (2) the voice-based agent through spoken prompts (achieved by the robot's voice with the robot concealed behind a black screen, out of the user's view), and (3) the LLM-powered social robot via spoken prompts, in a counterbalanced order. }
  \label{fig:teasor}
\end{figure}

%% file: figure_method.tex
\begin{figure}[t]
\centering
  \includegraphics[height=1.85in]{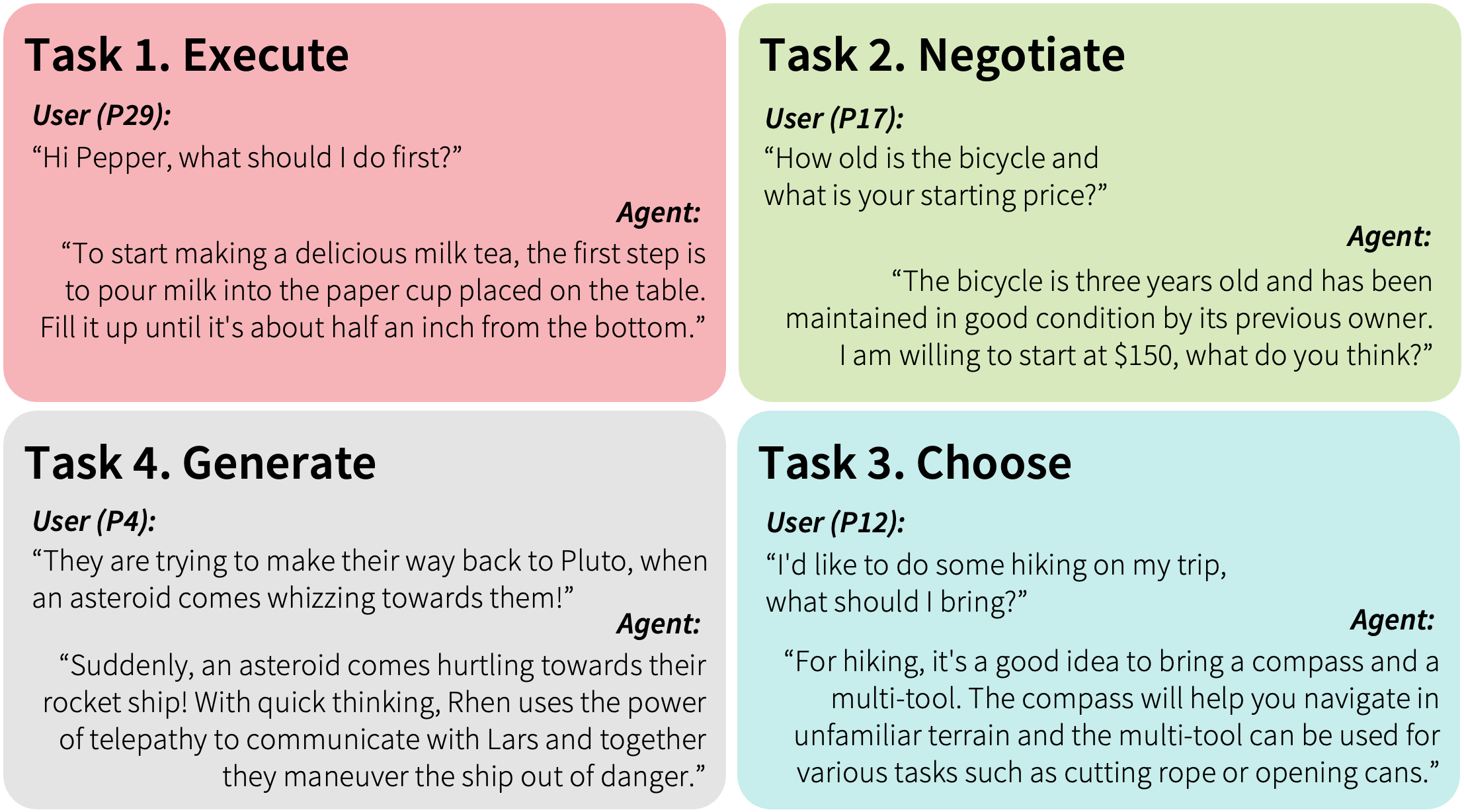}
  \caption{\textit{Interaction Examples per Each Task} --- Participants were assigned to one task among the four (\ie execute, negotiate, choose, and generate) and engaged with all three types of agents (\ie text, voice, and robot.) \textit{Top left to clockwise:} shows interaction examples of the four tasks. }
  \label{fig:method}
\end{figure}

%% file: test.tex
\begin{figure*}[!th]
  \includegraphics[width=\textwidth]{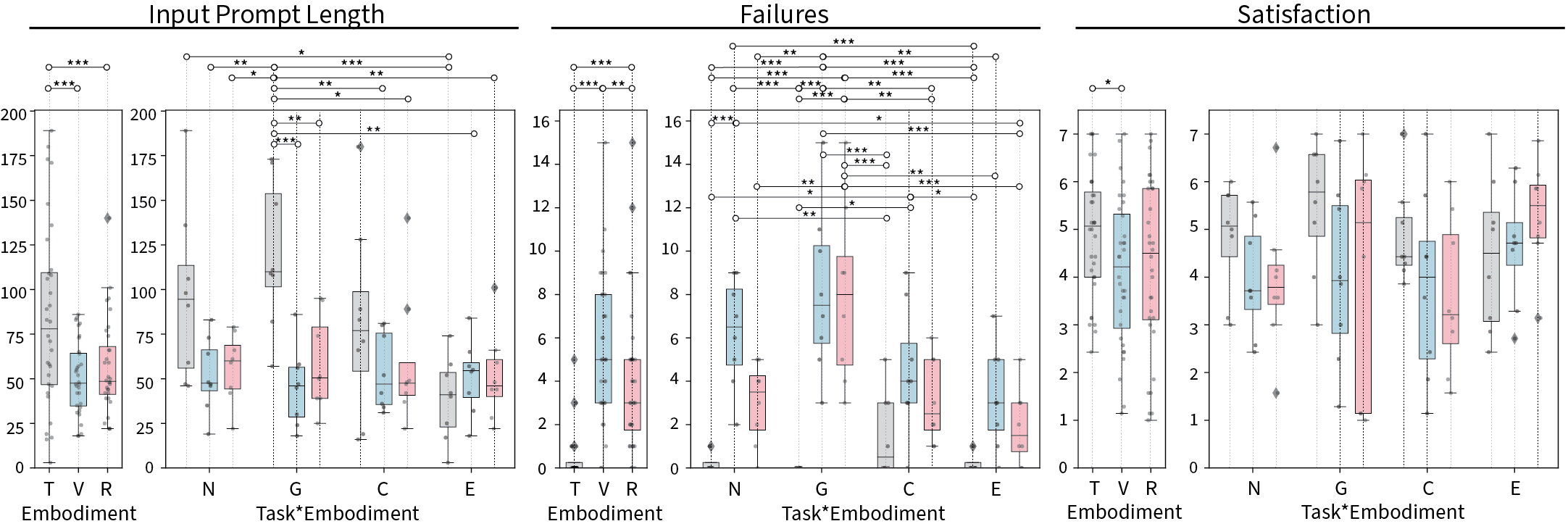}
   \vspace{-12pt}
  \caption{Boxplots with data points overlaid on user satisfaction, length of input prompts, and interaction failures. Embodiment: (T)ext, (V)oice, (R)obot. Tasks: (N)egotiate, (G)enerate, (C)hoose, (E)xecute. Horizontal lines indicate significant pairwise comparisons with Tukey HSD ($p < .05^{\ast}$, $p < .01^{\ast\ast}$, $p < .001^{\ast\ast\ast}$).}
  \label{fig:test}
\end{figure*}

%% file: figure_findings.tex
\begin{figure*}[!th]
  \includegraphics[width=\textwidth]{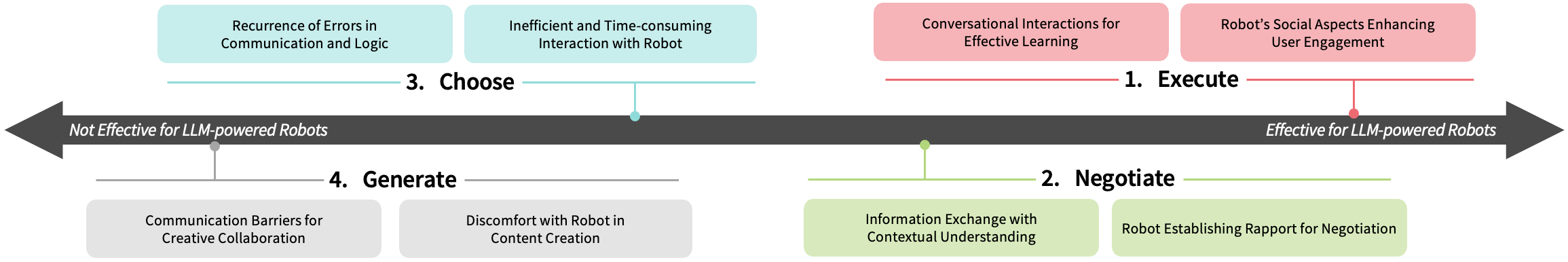}
  \caption{\textit{Summary of Qualitative Findings ---} Our findings indicate user preference for LLM-powered robots in the execution and negotiation tasks. These tasks necessitated the establishment of social relationships and rapport, and the robot's social aspects benefited from effective synergy with LLM capabilities. LLM-powered robots were less favored in the choice and generation tasks. In these cases, the robot's interaction medium and its social presence hindered optimal user performance. Additionally, a higher occurrence of technical communication errors contributed to participants' lower preference for robot agents.}
  \label{fig:findings}
\end{figure*}